\documentclass[conference]{IEEEtran}
\IEEEoverridecommandlockouts
\usepackage{cite}
\usepackage{amsmath,amssymb,amsfonts}
\usepackage{algorithmic}
\usepackage{graphicx}
\usepackage{textcomp}
\usepackage{colortbl}
\usepackage{color}
\usepackage[table]{xcolor}
\usepackage{booktabs}
\usepackage{hyperref}
\usepackage{multirow}

\def\BibTeX{{\rm B\kern-.05em{\sc i\kern-.025em b}\kern-.08em
    T\kern-.1667em\lower.7ex\hbox{E}\kern-.125emX}}

\DeclareRobustCommand{\IEEEauthorrefmark}[1]{%
  \smash{\textsuperscript{\footnotesize #1}}%
}
\renewcommand{\footnoterule}{%
  \kern -3pt%
  \hrule width 0.4\columnwidth height 0.4pt%
  \kern 2.6pt%
}

\begin{document}

\title{FAN-LoRA: A Fourier-Adaptive Nonlinear Low-Rank Adaptor for Medical Foundation Model Domain Adaptation}

\author{\IEEEauthorblockN{Ziquan Liu\IEEEauthorrefmark{1}, Zhewei Zhu\IEEEauthorrefmark{1}, and Xuyang Shi\IEEEauthorrefmark{1}$^*$}
\IEEEauthorblockA{\IEEEauthorrefmark{1}\textit{School of Information and Control Engineering, Southwest University of Science and Technology}\\
Mianyang, China}

\thanks{$^*$ Corresponding author. Email address: xuyangshi@swust.edu.cn (Xuyang Shi)}
}

\maketitle

\begin{abstract}
The advent of vision foundation models, notably the Segment Anything Model (SAM), has catalyzed significant advancements in natural image segmentation. However, their direct transfer to medical imaging remains severely bottlenecked by profound domain gaps, such as cross-modality and cross-center shifts. Existing Parameter-Efficient Fine-Tuning (PEFT) methods facilitate the adaptation of SAM to medical domains; nevertheless, they frequently suffer from performance degradation under severe distribution shifts. This vulnerability primarily stems from the implicit entanglement of heterogeneous frequency components within a shared low-rank subspace, which directly exacerbates sub-optimal structural alignment and localized boundary blurring. To overcome this representational bottleneck, we propose the Fourier-Adaptive Nonlinear Low-Rank Adaptor (FAN-LoRA), a novel frequency-decoupled fine-tuning architecture. FAN-LoRA explicitly separates the optimization space by employing a B-spline-driven low-pass branch for global structural alignment, synergistically coupled with a discrete Fourier high-pass branch for local textural compensation. Extensive experiments across three challenging cross-modality and cross-center benchmarks demonstrate that FAN-LoRA consistently outperforms state-of-the-art PEFT baselines. Compared to the strongest competitors, our method achieves consistent improvements in average Dice scores and notable reductions in boundary errors, while maintaining a compact module size without compromising computational efficiency.
\end{abstract}


\section{Introduction}
The advent of foundation models has catalyzed a paradigm shift in computer vision. Notably, the Segment Anything Model (SAM) \cite{kirillov2023segment}, pre-trained on the SA-1B dataset, has demonstrated remarkable zero-shot generalization in natural image segmentation. However, the direct transfer of SAM to medical image analysis remains severely bottlenecked by profound domain gaps \cite{mazurowski2023segment, huang2024segment}. Unlike natural images, medical modalities exhibit complex underlying physical mechanisms, subtle intensity variations, and highly heterogeneous lesion morphologies \cite{bougourzi2025recent, wang2022medical}. Consequently, applying pre-trained SAM directly to medical images frequently yields sub-optimal boundary delineation and significant performance degradation in low-contrast regions \cite{ji2024segment, huang2024segment, yao2024fnpc}.

To bridge this divide, contemporary research has pivoted to adapt SAM for specific medical domains. While large-scale full fine-tuning efforts, such as MedSAM \cite{MedSAM}, have established robust domain-specific baselines, their prohibitive computational costs hinder rapid deployment in specialized clinical settings. Consequently, Parameter-Efficient Fine-Tuning (PEFT) has emerged as the predominant approach \cite{lialin2023scaling, chen2022adaptformer, phuntsho2025adaptation, silva2025towards, jiang2025sr}. Methods leveraging Low-Rank Adaptation (LoRA) \cite{hu2022lora, samed} or domain-specific adapters \cite{wu2025medical} successfully inject task-specific knowledge with minimal parameter overhead. Nevertheless, current PEFT-adapted medical SAMs still suffer from severe performance degradation when confronting cross-center data or cross-modality domain shifts.

Existing PEFT methods implicitly entangle low-frequency structural drifts and high-frequency textural perturbations within a single low-rank subspace. This entanglement arises because domain discrepancies in medical imaging are inherently frequency-heterogeneous: low-frequency components dominate global intensity and anatomical shifts, while high-frequency components capture sensor-specific artifacts and lesion boundary details \cite{liu2021feddg, yang2020fda, li2025adaptive}. Consequently, gradient conflicts during optimization often lead to either over-smoothed representations or noise overfitting. This entanglement directly exacerbates the aforementioned sub-optimal boundary delineation and performance degradation in low-contrast regions.

To resolve this fundamental bottleneck, we propose the Fourier-Adaptive Nonlinear Low-Rank Adaptor (FAN-LoRA), a novel frequency-decoupled fine-tuning architecture. Diverging from conventional spatial-domain PEFT methods, we fundamentally re-model medical domain adaptation as a synergistic optimization of ``low-frequency structural alignment" and ``high-frequency residual compensation." 

Specifically, FAN-LoRA features a dual-branch architecture. For the low-pass branch, which governs global anatomical structure alignment, we draw inspiration from the continuous B-spline functions utilized in recent non-linear adapters \cite{dong2025aurora, cai2025loki}. Recognizing that the Fourier transform of B-spline basis functions exhibits a smooth low-pass response with polynomial decay in the frequency domain, we reconstruct the Adaptive Nonlinear Layer (ANL) with custom normalization and enhanced non-linear activations. This transforms the module into a highly efficient structural smoother, capturing the global intensity mapping between source and target domains on a low-frequency manifold. Concurrently, to compensate for the inevitable loss of local texture inherent in low-pass filtering, we introduce a complementary, explicit high-frequency branch. This branch constructs a sinusoidal projection using a fixed sinusoidal projection matrix, which approximates a subset of Fourier basis functions in the real space, effectively mapping spatial features into a discrete spectral subspace \cite{gao2024parameter}. Here, the model optimizes a highly parameter-efficient set of spectral parameters to capture target-specific high-frequency shifts before projecting back to the spatial domain. This cohesive design achieves explicit feature-level frequency decoupling, ensuring structural fidelity without compromising local detail.

The main contributions of this paper are summarized as follows:
\begin{itemize}
\item \textbf{Analytical Insight:} We identify and empirically analyze the performance bottleneck of frequency coupling in PEFT for cross-domain medical segmentation. We demonstrate that decoupling spatial gradients into B-spline-driven low-pass and Fourier-driven high-pass subspaces effectively mitigates optimization conflicts.
\item \textbf{Architectural Design:} We propose the FAN-LoRA architecture, which explicitly separates adaptation into a B-spline-driven low-pass branch for global structural alignment and a discrete Fourier high-pass branch for local textural compensation, rather than implicitly relying on shared parameterization.
\item \textbf{Empirical Effectiveness:} Extensive experiments across three challenging domain adaptation scenarios, including cross-modality and cross-center, demonstrate that FAN-LoRA achieves highly competitive performance compared to existing state-of-the-art PEFT methods in both accuracy and robustness.
\end{itemize}

\section{Related Work}
\subsection{Foundation Models in Medical Image Segmentation}
Prior to the advent of foundation models, the landscape of medical image segmentation was dominated by heavily supervised, task-specific architectures, most notably the nnU-Net framework \cite{isensee2021nnu}, which excelled through empirical pipeline optimization. The introduction of SAM \cite{kirillov2023segment} shifted the paradigm towards promptable, universal segmentation. However, the zero-shot application of SAM to medical modalities highlighted significant domain gaps, with recent comprehensive evaluations systematically characterizing its performance limitations on multi-modal medical data \cite{huang2024segment}.

To contextualize SAM for clinical use, comprehensive fine-tuning strategies were initially adopted. MedSAM \cite{MedSAM} aggregated a large-scale multi-modal dataset to train a universal medical baseline, demonstrating that domain-specific fine-tuning substantially improves segmentation accuracy. While these fully fine-tuned models possess strong generalized representations, their massive computational requirements render them inflexible for continuous adaptation to specific clinical centers or novel imaging protocols, highlighting the necessity for parameter-efficient alternatives.

\subsection{Parameter-Efficient Fine-Tuning and Adapters}
Parameter-efficient fine-tuning (PEFT) techniques aim to adapt large-scale pre-trained models to downstream tasks by updating only a minimal subset of parameters \cite{lialin2023scaling}. Within this paradigm, adapter-based methods inject lightweight trainable modules into frozen backbones. \cite{houlsby2019parameter} first introduced this concept in NLP, which was subsequently adapted to general vision tasks \cite{chen2022adaptformer}, and specifically tailored for medical SAM architectures via domain-specific adapters \cite{wu2025medical}. LoRA \cite{hu2022lora} offers an alternative approach by optimizing incremental low-rank weight matrices without modifying the base model's architecture. SAMed \cite{samed} pioneered the application of LoRA to SAM's image encoder for medical image segmentation, demonstrating that low-rank updates effectively inject domain-specific knowledge with minimal parameter overhead. Recent advancements further optimize this paradigm; for instance, DeLoRA \cite{bini2025delora} decouples magnitude and direction updates, while SHiRA \cite{bhardwaj2024shira} utilizes sparse high-rank adaptations for rapid fusion.

Despite their efficiency and structural improvements, standard linear PEFT methods face representational bottlenecks when modeling highly non-linear medical domain shifts. To address this, non-linear PEFT variants have been explored. Notably, AuroRA \cite{dong2025aurora} introduces an Adaptive Nonlinear Layer utilizing learnable B-spline basis functions, effectively expanding the approximation capacity of low-rank spaces. However, while AuroRA improves general gradient stability, it continues to operate within a unified parametric space. Consequently, it lacks an explicit mechanism to disentangle the conflicting high- and low-frequency gradients inherent in complex medical cross-modality shifts.

\subsection{Frequency-Domain Learning for Domain Adaptation}
Frequency-domain analysis has a rich history in mitigating domain shift in medical imaging. Classical unsupervised domain adaptation (UDA) methods, such as FDA \cite{yang2020fda}, demonstrated that swapping low-frequency Fourier amplitude spectra effectively transfers visual styles while preserving high-frequency semantic structures.

\begin{figure*}[ht]
\centering
\includegraphics[width=0.95\linewidth]{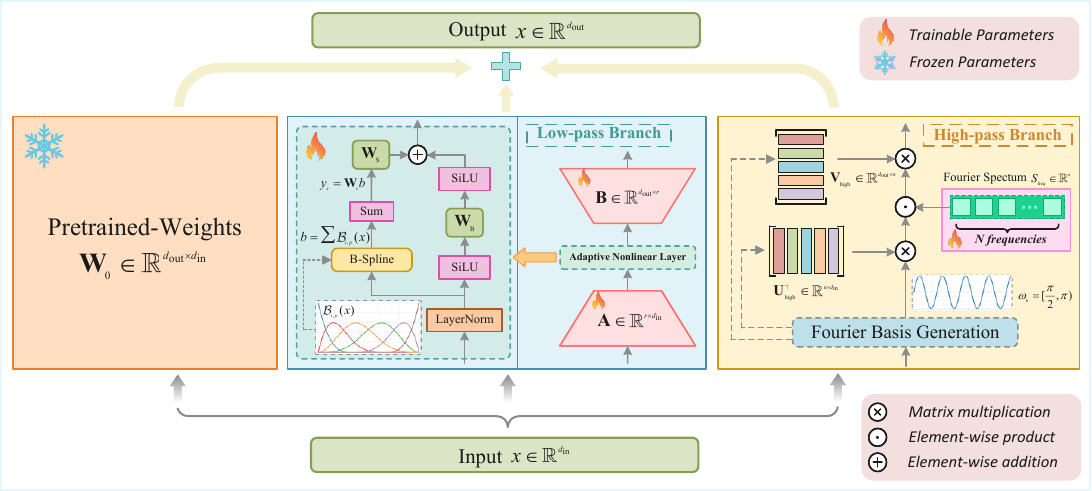}
\caption{Overall architecture of the proposed Fourier-Adaptive Nonlinear Low-Rank Adaptor (FAN-LoRA). The module fundamentally decouples spatial gradients by projecting input features into two parallel branches: a B-spline-driven Adaptive Nonlinear Layer (ANL) for low-frequency global structural alignment, and an explicit discrete Fourier branch for high-frequency local detail compensation.}
\label{fig:method}
\end{figure*}

More recently, frequency-aware mechanisms have been integrated into the PEFT paradigm. FourierFT \cite{gao2024parameter} treats spatial weight updates as spectral coefficients, optimizing a sparse set of Fourier parameters to maximize efficiency in general vision and language tasks. In the medical domain, AdaptFRCNet~\cite{he2025adaptfrcnet} incorporates frequency-domain consistency regularization within a semi-supervised PEFT framework, leveraging frequency cues to enhance segmentation performance with limited labeled data. Similarly, FreqFiT~\cite{ly2025frequency} introduces a frequency-based fine-tuning module inserted between ViT blocks, capturing global anatomical dependencies that spatial-domain PEFT methods often overlook. However, these existing frequency-based adapters apply spectral operations without structurally separating the optimization of low-frequency structural priors from high-frequency textural details.

Our proposed FAN-LoRA departs from this paradigm by enforcing explicit feature-level frequency decoupling—utilizing continuous splines strictly for low-frequency structural alignment and discrete Fourier transforms strictly for high-frequency residual compensation.

\section{Methodology}
To address the domain shift issues encountered by medical foundation models when applied to varying modalities and clinical center data, we propose a parameter-efficient fine-tuning module based on frequency decoupling, the Fourier-Adaptive Nonlinear Low-Rank Adaptor (FAN-LoRA), as shown in Fig.~\ref{fig:method}. The core idea is that domain shifts in medical images manifest in the frequency domain as a mixture of low-frequency structure changes and high-frequency detail variations. Traditional linear adapters tend to lose high-frequency details or overfit low-frequency noise during optimization. Therefore, FAN-LoRA introduces parallel branches comprising a B-spline Adaptive Nonlinear Layer (ANL) with low-pass characteristics and an explicit lightweight Fourier high-pass filter.

\subsection{Overall Architecture}
Given the frozen weight matrix $\mathbf{W}_0 \in \mathbb{R}^{d_{\text{out}} \times d_\text{in}}$ of the pre-trained model and the input feature $x \in \mathbb{R}^{d_\text{in}}$, the forward propagation of FAN-LoRA consists of the base linear mapping, a low-pass adaptation branch and a high-pass adaptation branch:
\begin{equation}
y = \mathbf{W}_0 x + H_\text{low}(x) + H_\text{high}(x)
\end{equation}
where $H_\text{low}(x)$ and  $H_\text{high}(x)$ denote the low-frequency structural adaptation and high-frequency residual compensation, respectively.

\subsection{Low-Pass Adaptation via B-Spline ANL}
To capture global, smooth domain distribution changes, we introduce the ANL in the low-rank space. We first project the features to a low-rank dimension $r$ using the down-projection matrix $\textbf{A} \in \mathbb{R}^{r \times d_\text{in}}$, process them through the ANL module, and then map them back to the original space via the up-projection matrix $\mathbf{B} \in \mathbb{R}^{d_\text{out} \times r}$:
\begin{equation}
H_\text{low}(x) = \frac{\alpha}{r} \textbf{B} \cdot \Phi_{\text{ANL}}(\textbf{A} \cdot \mathcal{D}(x))
\end{equation}
where $r$ is the intrinsic rank, and $\alpha$ is a scaling factor controlling the adaptation magnitude. The stochastic dropout $\mathcal{D}(\cdot)$, inherited from AuroRA~\cite{dong2025aurora}, is applied before the low-rank projection to regularize the bottleneck subspace and mitigate overfitting of the B-spline basis functions to source-domain noise patterns. $\Phi_{\text{ANL}}(\cdot)$ represents the ANL composed of a SiLU activation and a B-spline residual branch. 

The ANL module contains a basic non-linear transformation and a residual branch based on B-splines. For the input feature $z = \textbf{A} x$, we apply LayerNorm followed by a non-linear transformation and a spline-based residual mapping. In practice, we adopt a spline order $p=3$ with a fixed grid size of $G=5$, uniformly distributed over the normalized input range  \cite{dong2025aurora}. Its core lies in the aggregated output of the B-splines:
\begin{equation}
\text{Spline}(z) = \sum_{i=1}^G w_i \mathcal{B}_{i, p}(\text{LayerNorm}(z))
\end{equation}
where $\mathcal{B}_{i,p}(\cdot)$ is the B-spline basis of degree $p$, and $w_i$ denotes the learnable spline weights.

From a signal processing perspective, the Fourier transform of a $p$-degree B-spline basis function is proportional to $\text{sinc}^{p+1}(f)$, whose magnitude exhibits increasingly rapid decay at high frequencies as the spline order increases~\cite{cordero2020generalized}. This property implies that B-spline basis functions act as smooth low-pass filters with a polynomially decaying frequency response~\cite{unser1993family}.  Consequently, B-spline-based interpolation can be interpreted as a low-pass filtering process that suppresses high-frequency oscillations and noise while preserving the underlying low-frequency structure~\cite{panda2001least}. Such smoothing behavior has been widely exploited in signal and image processing to achieve stable approximation of noisy data~\cite{panda2001least}.  In the context of medical imaging, this property is particularly beneficial, as it promotes smooth structural alignment across domains and facilitates the learning of consistent global anatomical representations.

\subsection{High-Pass Compensation via Lightweight FourierFT}
While the low-pass branch effectively aligns the global domain distribution, it tends to smooth out minute textures and sharp boundaries of medical lesions (e.g., spiculations on tumor margins). To address this, we design a lightweight high-pass Fourier branch $H_\text{high}(x)$ for frequency compensation.

We explicitly define the high-frequency basis as sine waves in the discrete space. Assuming the number of sampled high frequencies is $n=16$ (default; see ablation in Table~\ref{tab:prostate_hyperparameters}), we uniformly sample the high-frequency band $\omega \in [\frac{\pi}{2}, \pi)$. By interpreting the permutation-invariant channel dimensions as a discrete sequential signal, we project the features into an implicit spectral subspace—a technique conceptually aligned with implicit neural representations. Since the Nyquist frequency of discrete signals is $\pi$, this half-open interval heuristically emphasizes higher-frequency components while strictly avoiding the singularity at $\pi$, ensuring no spectral parameters are wasted. The high-frequency projection matrices $\mathbf{U}_\text{high} \in \mathbb{R}^{d_\text{in} \times n}$ and $\mathbf{V}_\text{high} \in \mathbb{R}^{d_\text{out} \times n}$ are defined as:
\begin{equation}
\mathbf{U}_\text{high}^{(i, j)} = \frac{1}{\sqrt{d_\text{in}}} \sin(i \cdot \omega_j), \quad \mathbf{V}_\text{high}^{(k, j)} = \frac{1}{\sqrt{d_\text{out}}} \sin(k \cdot \omega_j)
\end{equation}
where $i \in \{1, \dots, d_\text{in}\}$ and $k \in \{1, \dots, d_\text{out}\}$ are feature dimension indices, and $\omega_j$ is the $j$-th sampled high frequency. These two matrices are generated dynamically and require no gradient calculations.

The forward process of the high-pass branch is implemented via a learnable high-frequency spectral modulation vector $S_\text{freq} \in \mathbb{R}^{n}$:
\begin{equation}
H_\text{high}(x) = \mathbf{V}_\text{high} ( S_\text{freq} \odot(\mathbf{U}_\text{high}^\top x))
\end{equation}
This branch projects features into a high-frequency subspace via fixed sinusoidal bases, modulates them with a learnable spectral vector, and reconstructs the residual in the original feature space. Compared to conventional approaches, this design avoids explicitly constructing large dense weight matrices in the high-frequency space, achieving extreme parameter efficiency (requiring only $n$ parameters).

\begin{figure}[h]
\centering
\includegraphics[width=0.95\linewidth]{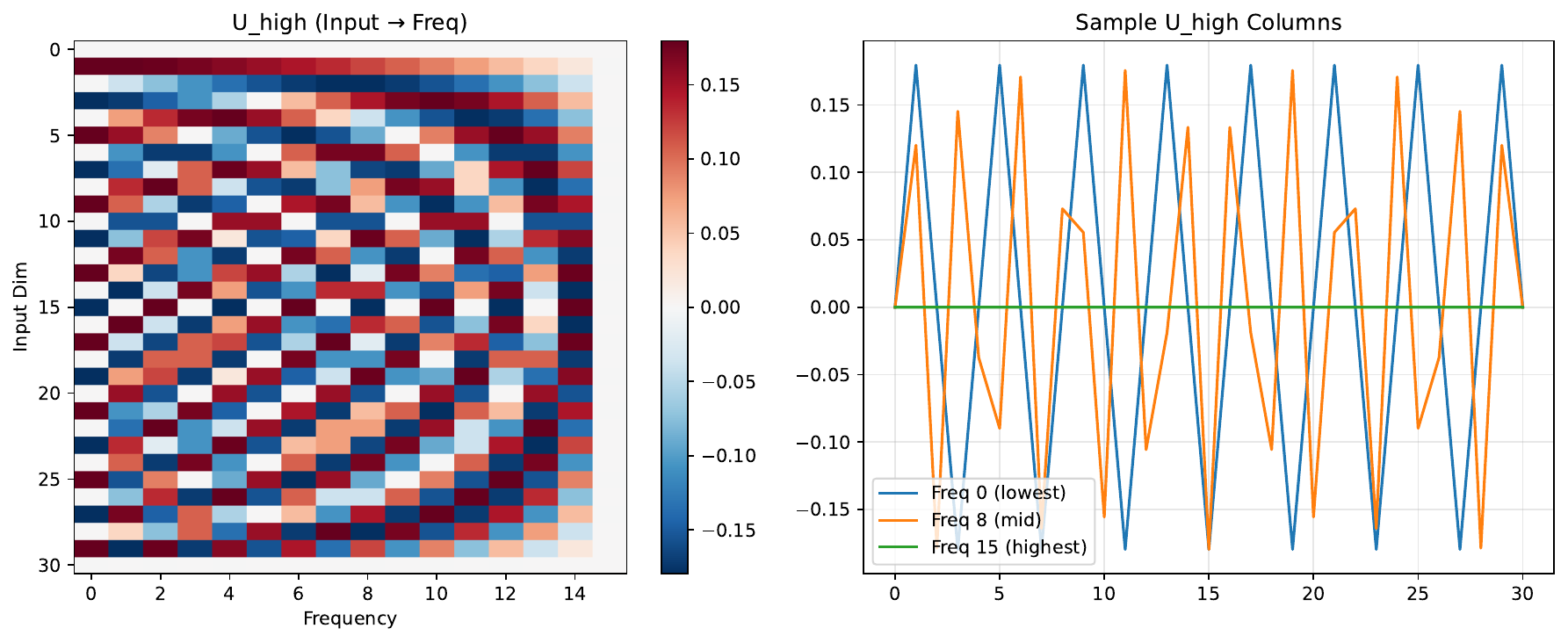}
\caption{Visualization of the discrete Fourier projection basis matrix $\mathbf{U}_{\text{high}}$. The columns are generated via sinusoidal sampling along discrete integer dimensions, capturing distinct high-frequency components to effectively extract texture residuals without excessive parameter overhead.}
\label{fig:fourier}
\end{figure}

The column vectors of the Fourier projection basis $\mathbf{U}_\text{high}$ are generated by sinusoidal sampling along discrete integer dimensions, as shown in Fig.~\ref{fig:fourier}. The lowest frequency component ($0.5\pi$) exhibits a regular sinusoidal pattern due to its integer period; the mid-frequency components suffer from discrete sampling mismatch due to their non-integer periods, resulting in natural waveform distortion. 

The effectiveness of FAN-LoRA in MedSAM domain adaptation stems from its gradient decoupling in the frequency domain. In traditional LoRA fine-tuning, the updating of the weight matrix must simultaneously accommodate large-scale pixel drifts (low frequency) and local texture mutations (high frequency). This often leads to optimization dilemmas in medical multi-modality data (e.g., transferring from CT to MRI). FAN-LoRA explicitly resolves this dilemma: the B-spline branch smoothly absorbs large-scale low-frequency domain shifts, while the Fourier branch precisely compensates for high-frequency structural boundaries, achieving robust and decoupled domain adaptation. 

\section{Experiments}

\subsection{Datasets}
We evaluate FAN-LoRA on three representative medical image segmentation benchmarks under cross-modality and cross-center settings, as shown in Table~\ref{tab:datasets}. Specifically, MM-WHS 2017 \cite{zhuang2019mmwhs} is used for cross-modality cardiac segmentation, with MR as the source domain and CT as the target domain. Promise 12 \cite{litjens2014promise} and NCI-ISBI \cite{bloch2015nciisbi} are employed for cross-center prostate segmentation under different scanner conditions. FLARE 22 \cite{ma2023flare} and CHAOS \cite{kavur2021chaos} are adopted for cross-modality abdominal organ segmentation, where CT serves as the source domain and MRI as the target domain. These datasets cover diverse anatomical structures and domain shift patterns, enabling a comprehensive evaluation of adaptation performance.

\begin{table}[h]
\centering
\caption{Summary of datasets and domain adaptation settings.}
\label{tab:datasets}
\resizebox{1.0\linewidth}{!}{
\begin{tabular}{lccc}
\toprule
Source & Target & Modality & Anatomy \\
\midrule
MM-WHS 2017 MR & MM-WHS 2017 CT & MR $\rightarrow$ CT & Heart \\
Promise 12 & NCI-ISBI & MRI (1.5T/3T) & Prostate \\
FLARE 22 & CHAOS & CT $\rightarrow$ MRI & Abdomen \\
\bottomrule
\end{tabular}
}
\end{table}

\subsection{Implementation Details}
All experiments were conducted on a single NVIDIA Tesla V100 GPU (32GB). We adopted the officially pre-trained MedSAM model as our backbone architecture. During fine-tuning, all parameters of the pre-trained backbone were kept frozen, and gradients were exclusively computed for the newly inserted FAN-LoRA modules. For data preprocessing, input images and their corresponding segmentation masks were resized to a uniform spatial resolution of $1024 \times 1024$ pixels to meet the model's input constraints. The network was optimized for 100 epochs using the AdamW optimizer with a batch size of 2. The initial learning rate was set to $1\times 10^{-4}$ for the LoRA parameters, coupled with a mild weight decay of $1\times 10^{-3}$ to regularize the newly added adapters. The training objective was formulated as a linear combination of Dice loss and Binary Cross-Entropy (BCE) loss.

Supervised fine-tuning was performed exclusively on the source domain using identical data splits across all methods to ensure fair comparison. All experiments used the official MedSAM ViT-B/16 pre-trained weights with bounding box prompts derived from the masks (expanded by a random margin of 0-50 pixels to simulate human interactions), following standard evaluation protocols.

\subsection{Experimental Results}

\textbf{Comparison with Existing Adapters:} We quantitatively compared FAN-LoRA with mainstream parameter-efficient fine-tuning methods on the four target domain datasets. These methods include LoRA \cite{hu2022lora}, its variants such as DeLoRA \cite{bini2025delora} and SHiRA \cite{bhardwaj2024shira}, nonlinear adapter AuroRA \cite{dong2025aurora}, and the frequency-based method FourierFT \cite{gao2024parameter} and FreqFiT~\cite{ly2025frequency}, covering a variety of design paradigms, including linear, nonlinear, and frequency-domain adaptive. The evaluation metrics primarily include the Dice Similarity Coefficient (DSC) and the 95\% Hausdorff Distance (HD95). The quantitative results are presented in Tables ~\ref{tab:chaos_comparison}, \ref{tab:prostate_comparison}, and \ref{tab:mmwhs_comparison}. FAN-LoRA achieves the best performance among all compared PEFT methods on the CHAOS, NCI-ISBI, and MM-WHS target domains, delivering consistent DSC gains (e.g., an absolute +1.0\% over the strongest spectral baseline on MM-WHS) and achieving the lowest HD95 across all examined organs.

\begin{table}[ht]
\centering
\caption{Quantitative comparison of different PEFT methods on the CHAOS target domain (MRI)}
\label{tab:chaos_comparison}
\resizebox{1.0\linewidth}{!}{
\begin{tabular}{l|cccccc}
\toprule
Method & \multicolumn{6}{c}{CHAOS} \\
& Liver & R.kidney & L.kidney & Spleen & Avg. & HD95 $\downarrow$ \\
\midrule
Zero-shot   & 90.9 & 90.5 & 89.8 & 93.6 & 91.2 & 5.03 \\
Fine-tuning   & 72.6 &  87.5& 84.4& 88& 83.1 & 11.21 \\
LoRA    & 94.3& 91.6& 91.5& 94.6& 93 & 2.84 \\
FourierFT    & 94.7& 92.5& 91.4& 94.5& 93.3 & 2.71 \\
DeLoRA    & 94.4& 92.9& 91.6& 94.2& 93.3 & 2.66 \\
SHiRA    & 94.5& 92.3& 91& 94.3& 93 & 2.95 \\
FreqFiT & \textbf{94.8} & 92.6& 90.2& 93.9& 92.9 & 3.28 \\
AuroRA    & 94.1& 93& 92.1& 94.4& 93.4 & 3.04 \\
\textbf{FAN-LoRA (Ours)} & 94.5& \textbf{93.4}& \textbf{92.7}& \textbf{94.8}& \textbf{93.9} & \textbf{2.32} \\
\bottomrule
\end{tabular}
}
\end{table}

As shown in Table~\ref{tab:chaos_comparison}, compared with AuroRA, which relies solely on non-linear mapping, FAN-LoRA achieves better boundary accuracy, indicating that non-linearity alone is insufficient to recover high-frequency details. Similarly, although FourierFT improves over LoRA by introducing spectral modeling, its lack of structural constraints leads to suboptimal global consistency. The superior performance of FAN-LoRA suggests that explicitly decoupling low-frequency structural alignment and high-frequency detail compensation is crucial for cross-modality adaptation from CT to MRI, where both intensity distribution and boundary characteristics differ consistently.

Note that the zero-shot and full fine-tuning baselines exhibit notably lower Dice on CHAOS compared to LoRA; this is attributed to the severe CT→MRI intensity gap causing overfitting in full-parameter tuning, which our frequency-decoupled design effectively mitigates.
\begin{table}[ht]
\centering
\caption{Quantitative comparison of different PEFT methods on the NCI-ISBI target domain across 1.5T and 3T scanners}
\label{tab:prostate_comparison}
\resizebox{1.0\linewidth}{!}{
\begin{tabular}{l|cccc|cccc}
\toprule
Method & \multicolumn{4}{c|}{NCI-ISBI 1.5T} & \multicolumn{4}{c}{NCI-ISBI 3T} \\
& PZ & CG & Avg. & HD95 $\downarrow$ & PZ & CG & Avg. & HD95 $\downarrow$ \\
\midrule
Zero-shot   & 67.2& 87.7& 77.5 & 37.68 & 66.4& 89.9& 78.2 & 31.52\\
Fine-tuning   & 65.8& 90.2& 78 & 34.3 & 58.9& 91.7& 75.3 & 33.69 \\
LoRA    & 66.3& \textbf{93.5}& 79.9 & 30.04 & 60.2& 93.1& 76.7 & 31.99 \\
FourierFT    & 68.2& 92.7& 80.5 & 27.87 & 62.8& 93.1& 78 & 27.51\\
DeLoRA    & 65.7& 93.2& 79.5 & 30.4 & 65.5& 92.8& 79.2 & 27.12 \\
SHiRA    & 67.4& 91.9& 79.7 & 28.91 & 59.1& 92.9& 76 & 32.93 \\
FreqFiT  & 65.1 & 92.6 & 78.9 & 27.99 & 61.8 & 92 & 76.9 & 28.34 \\
AuroRA    & 66.8& 91.4& 79.1 & 29.29 & 62.6& 91.8& 77.2& 29.68\\
\textbf{FAN-LoRA (Ours)} & \textbf{68.4}& 93.3& \textbf{80.9} & \textbf{26.96} & \textbf{66.4}& \textbf{92.8}& \textbf{79.6} & \textbf{26.48} \\
\bottomrule
\end{tabular}
}
\end{table}

Table~\ref{tab:prostate_comparison} evaluates robustness under cross-center variations, where domain shifts mainly manifest as intensity distribution changes and acquisition noise. FAN-LoRA consistently achieves the highest Dice scores and the lowest HD95 on both 1.5T and 3T settings. Notably, its improvement over LoRA is more pronounced in HD95 than in Dice, indicating that FAN-LoRA primarily enhances boundary precision rather than coarse region overlap. This aligns with our design motivation: the high-frequency branch explicitly targets boundary refinement, which is critical under scanner-induced noise. Compared with FourierFT, FAN-LoRA also exhibits better stability across different magnetic field strengths, suggesting that simply introducing frequency modeling is insufficient without the structural guidance of the low-pass branch. Similarly, FreqFiT, another frequency-oriented adapter, lags behind FAN-LoRA, particularly in boundary accuracy, further confirming that high-frequency optimization becomes unstable in the absence of explicit frequency decoupling.

\begin{table}[ht]
\centering
\caption{Quantitative comparison of different PEFT methods on the MM-WHS 2017 CT target domain}
\label{tab:mmwhs_comparison}
\resizebox{1.0\linewidth}{!}{
\begin{tabular}{l|cccccc}
\toprule
Method & \multicolumn{6}{c}{MM-WHS 17 CT} \\
& LAC & LVC & MYO & AA & Avg. & HD95 $\downarrow$ \\
\midrule
Zero-shot   & 68.5& 91.3& 87.2& 90.7& 84.4 & 30.07 \\
Fine-tuning   & 70.2& 91.9& 87.3& 91.3& 85.2 & 28.79 \\
LoRA    & 74.1& 91.8& 88.5& 91& 86.4 & 27.84 \\
FourierFT    & 86.3& 91.9& 89.3& 91.4& 89.7 & 12.97 \\
DeLoRA    & 73.8& 91.6& 88.3& 91.2& 86.5 & 26.95 \\
SHiRA    & 81.7& 91.6& 88.3& 91.2& 88.2 & 20.83 \\
FreqFiT  & 85.3 & 92.4 & 87.8 & 89.6 & 88.8 & 19.44 \\
AuroRA    & 79.9& 91.8& 85.7& 90.7& 87 & 20.78 \\
\textbf{FAN-LoRA (Ours)} & \textbf{86.4}& \textbf{93}& \textbf{90.4}& \textbf{92.8}& \textbf{90.7} & \textbf{12.47} \\
\bottomrule
\end{tabular}
}
\end{table}

\begin{table*}[ht]
\centering
\caption{Performance retention evaluation of different PEFT methods on the FLARE 22 source domain (CT)}
\label{tab:source_comparison_1}
\resizebox{1.0\linewidth}{!}{
\begin{tabular}{l|cccccccccccccc}
\toprule
Method & \multicolumn{14}{c}{FLARE 22} \\
 & Liver & R.kidney & Spleen & Pancreas & Aorta & IVC & RAG & LAG & Gallbladder & Esophagus & Stomach & L.kidney & Avg. & HD95 $\downarrow$ \\
\midrule
Zero-shot & 96 & 95.8 & 96.2 & 83.2 & 94.3 & 92 & 56.2 & 67.2 & 88.8 & 78.3 & 90.7 & 68.9 & 83.97 & 11.02 \\
Fine-tuning & 96.2 & 96.4 & 97.4 & 83.5 & 95.9 & 93.4 & 77.2 & 77.3 & 90.3 & 88.7 & 92.9 & 80.3 & 89.13 & 4.88 \\
LoRA & 96.4 & 96.6 & 96.7 & 84.3 & 94.7 & 94.2 & 68.8 & 67.6 & 91 & 89.1 & 93.4 & 74 & 87.23 & 7.1 \\
FourierFT & 95.9 & 95.8 & 95.5 & \textbf{87.4} & 93.6 & 93.9 & 71.7 & 74.4 & 90.6 & 88.7 & 92.3 & 78.9 & 88.23 & 5.32 \\
DeLoRA  & 96.3 & 96.4 & 96.4 & 86.4 & 94.7 & \textbf{94.4} & 75.2 & 76 & 90.9 & 88.5 & 92.9 & 81 & 89.09 & \textbf{4.34} \\
SHiRA  & 95.8 & 96.1 & 96.6 & 84.3 & 94.3 & 93 & 64.9 & 70.8 & 89.9 & 86.7 & 91 & 77 & 86.7 & 6.54 \\
FreqFiT  & 94.1 & 95.8 & 96.9 & 84.5 & 95.2 & 92.7 & 65.1 & 71.6 & 88.3 & 86.6 & 91.2 & 78.6 & 86.7 & 6.98 \\
AuroRA  & 95.2 & 96.0 & 95.8 & 83.4 & 94.1 & 92.9 & 72.9 & 74.8 & 90.4 & 89.4 & 92.3 & 77.5 & 87.89 & 5.92\\
\textbf{FAN-LoRA (Ours)} & \textbf{97.1} & \textbf{96.8} & \textbf{97.5} & 85.7 & \textbf{96.0} & 94.3 & \textbf{78.2} & \textbf{81.1} & \textbf{92.1} & \textbf{90.1} & \textbf{94.0} & \textbf{83.1} & \textbf{90.50}  & 5.56 \\
\bottomrule
\end{tabular}
}
\end{table*}

\begin{table}[ht]
\centering
\caption{Performance retention evaluation of different PEFT methods on the Promise 12 and MM-WHS 2017 MR source domains}
\label{tab:source_comparison_2}
\resizebox{1.0\linewidth}{!}{
\begin{tabular}{l|cc|cccccc}
\toprule
Method & \multicolumn{2}{c|}{Promise 12} & \multicolumn{6}{c}{MM-WHS 17 MR} \\
& prostate & HD95 $\downarrow$ & LAC & LVC & MYO & AA & Avg. & HD95 $\downarrow$ \\
\midrule
Zero-shot   & 92.3& 4.58& 52.5 & 80.8 & 82.9 & 73.3 & 72.38 & 42.15 \\
Fine-tuning   & 94.2& 3.12& 84.5 & 83.2 & 88.7 & 85.2 & 85.4 & 20.03 \\
LoRA    & 94& 3.45& 80.2 & 83.2 & 87.9 & \textbf{82.9} & 83.6 & 20.55 \\
FourierFT    & 94.1& 2.85& 81.1 & 82.8 & 87.2 & 82.1 & 83.37 & 23.46 \\
DeLoRA    & 94.1& 3.11& 85.2 & 82.8 & 88.4 & 79.3 & 83.9 & 21.28 \\
SHiRA    & 93.9& 3.07& 80.6 & 82.9 & 87.6 & 78.7 & 82.5 & 22.64 \\
FreqFiT    & 94.92& 2.79& 82.8 & 83.1 & 86.9 & 79.7 & 83.1 & 23.23 \\
AuroRA    & 93.8& 3.16& 81.7 & 82.9 & 87.5 & 81 & 83.3 & 28.17 \\
\textbf{FAN-LoRA (Ours)} & \textbf{95}& \textbf{1.89}& \textbf{84.7} & \textbf{83.6} & \textbf{88.5} & 82.8 & \textbf{84.9} & \textbf{19.63} \\
\bottomrule
\end{tabular}
}
\end{table}

In the challenging MM-WHS cross-modality setting, FAN-LoRA demonstrates a substantial performance gain over all baselines. Unlike CHAOS, this task involves more complex anatomical structures and stronger modality gaps, making it particularly sensitive to structural misalignment, as shown in Table~\ref{tab:mmwhs_comparison}. Methods such as SHiRA and AuroRA improve over LoRA but still struggle with consistent multi-organ alignment, as reflected by their lower average Dice scores. Among them, SHiRA achieves a competitive Dice of 88.2 yet exhibits a relatively high HD95 of 20.83. This suggests that while its sparse high-rank adapter updates are sufficient for coarse region coverage, they provide insufficient gradient density along organ boundaries, especially at the myocardial-blood pool interface, which leads to degraded boundary precision. FourierFT consistently boosts performance by capturing spectral information; however, its lack of explicit structural modeling limits further gains. FAN-LoRA achieves the best performance by jointly optimizing global structure and local details. The low-pass branch ensures stable anatomical alignment across modalities, while the high-pass branch refines organ boundaries. This synergy is especially beneficial for complex cardiac structures, where both global topology and local boundary precision are critical.

\textbf{Performance Retention on Source Domains:} Tables~\ref{tab:source_comparison_1} and~\ref{tab:source_comparison_2} demonstrate that FAN-LoRA not only adapts effectively to target domains but also preserves performance on source domains, avoiding catastrophic forgetting. Interestingly, FAN-LoRA even slightly improves performance over full fine-tuning in some cases (e.g., FLARE 22), suggesting that the frequency-decoupled design acts as an implicit regularizer. By restricting low-frequency and high-frequency updates into separate subspaces, the model avoids overfitting to domain-specific noise while preserving general anatomical priors. This property is particularly desirable in clinical applications, where maintaining robustness across multiple domains is often more critical than optimizing for a single dataset.

\begin{table}[h]
\centering
\caption{Comparison of trainable parameters for different PEFT adapters when integrated into the MedSAM image encoder.}
\resizebox{0.6\linewidth}{!}{
\begin{tabular}{l|c}
\toprule
Method & Trainable Parameters \\
\midrule
LoRA & 454,656 \\
FourierFT & 24,000 \\
DeLoRA & 442,392 \\
SHiRA & 442,368 \\
FreqFiT & 592128 \\
AuroRA & 111,072 \\
\textbf{FAN-LoRA (Ours)} & 111,552 \\
\bottomrule
\end{tabular}
}
\label{tab:params}
\end{table}

\textbf{Parameter Efficiency Analysis:} As shown in Table~\ref{tab:params}, LoRA-based methods introduce over 440K trainable parameters, while FourierFT is highly efficient but limited in structural modeling. AuroRA reduces the parameter count to 111K, and FAN‑LoRA maintains a nearly identical footprint of approximately 111.5K parameters despite its additional lightweight Fourier branch. This makes FAN‑LoRA roughly four times more parameter-efficient than standard LoRA (over 450K parameters) while delivering superior segmentation accuracy and boundary precision, demonstrating a clearly favorable efficiency–performance trade-off. In contrast, FreqFiT incurs an even larger parameter budget (592K) than LoRA, yet fails to match FAN‑LoRA in either metric, further confirming that a larger parameter count without explicit frequency decoupling does not translate into better domain adaptation.

\begin{figure}[ht]
\centering
\includegraphics[width=0.95\linewidth]{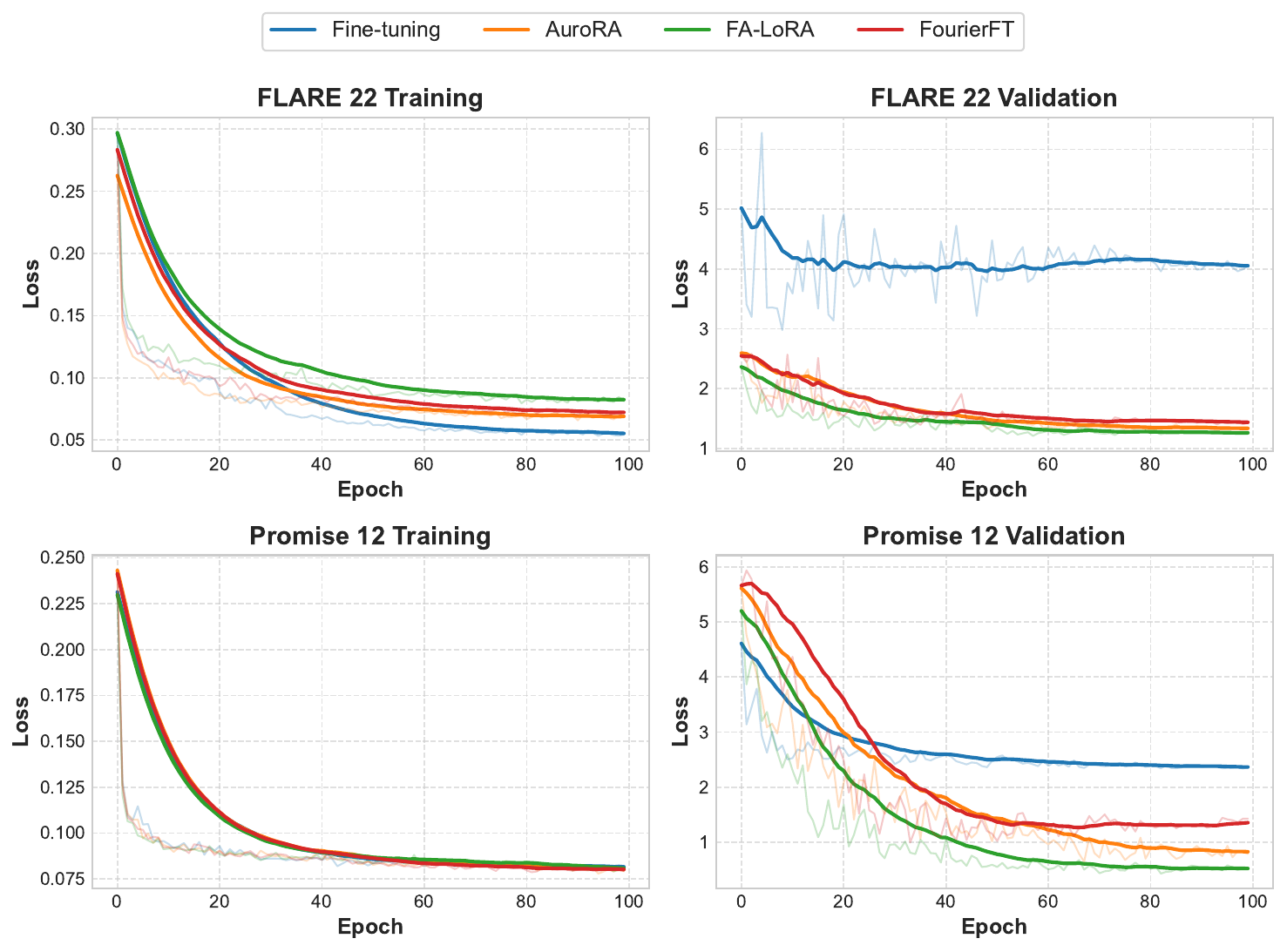}
\caption{Training and validation loss curves on the FLARE 22 and Promise 12 datasets, initialized with SAM pre-trained weights. Four methods are compared: Fine-tuning, AuroRA, FourierFT, and FAN‑LoRA. On both datasets, Fine-tune exhibits signs of overfitting. In contrast, compared with AuroRA and FourierFT, FAN‑LoRA demonstrates a smoother convergence trajectory and exhibits less overfitting.}
\label{fig:loss_curve}
\end{figure}

\textbf{Training Dynamics and Loss Curve Analysis:} As illustrated in Fig.~\ref{fig:loss_curve}, FAN-LoRA exhibits a smoother and more stable convergence compared to AuroRA and FourierFT. Notably, the validation loss of FAN-LoRA shows fewer oscillations, indicating reduced overfitting to high-frequency noise. This behavior can be attributed to the explicit frequency decoupling mechanism. In contrast, AuroRA operates in a unified non-linear space, where gradients from different frequency components may interfere, while FourierFT lacks sufficient constraints on structural consistency. By separating optimization into low- and high-frequency subspaces, FAN-LoRA effectively stabilizes gradient updates, leading to improved generalization.

\begin{figure*}[ht]
\centering
\includegraphics[width=0.95\linewidth]{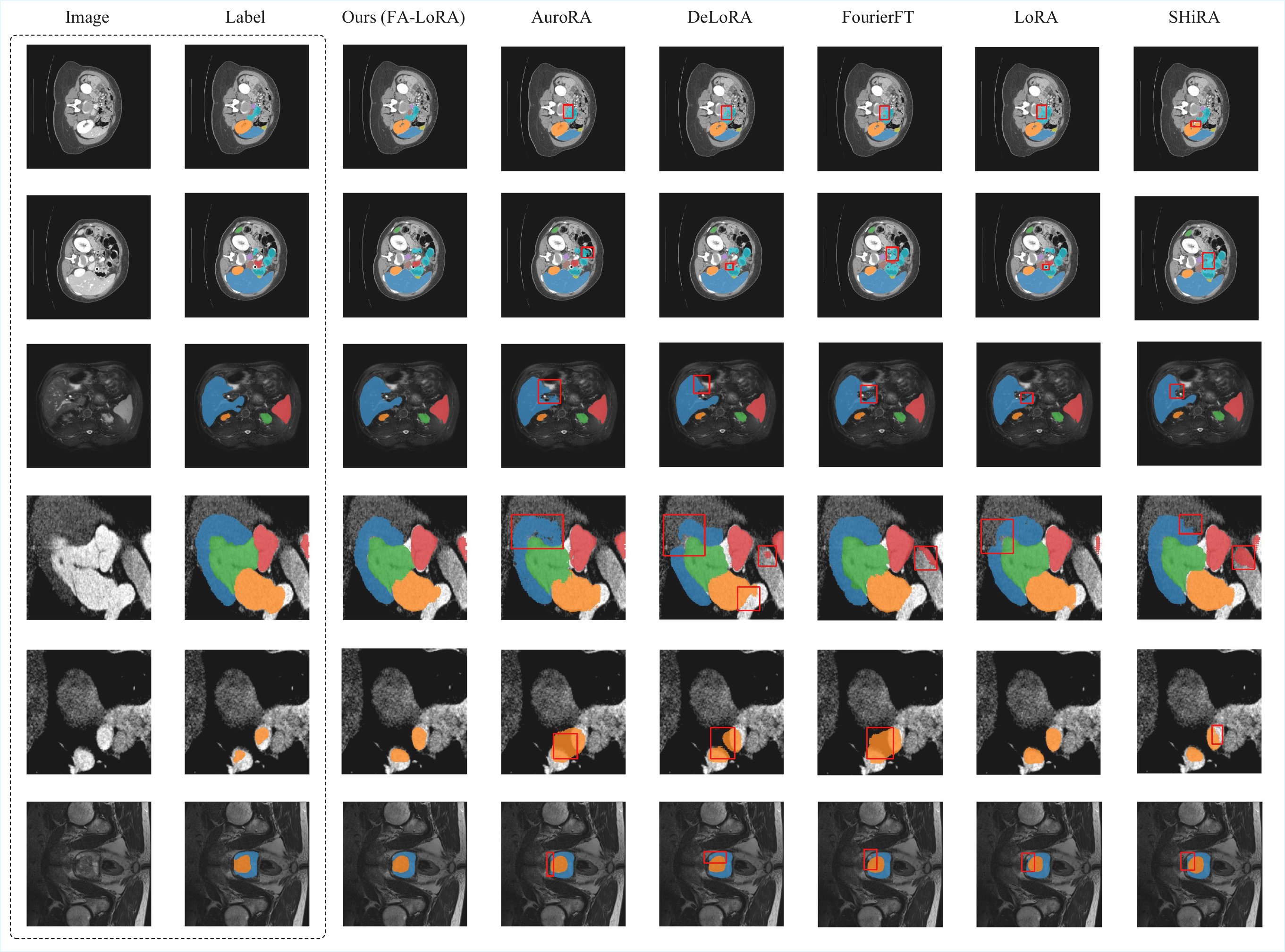}
\caption{Qualitative comparison of different domain adaptation methods. From left to right: input image, ground truth, FAN-LoRA (ours), AuroRA, DeLoRA, FourierFT, LoRA, and SHiRA. FAN-LoRA delivers sharper boundaries and better-preserved fine structures under domain shift, while competing methods tend to over-smooth or produce noisy artifacts. Red boxes highlight representative errors, such as missed structures and false positives.}
\label{fig:Visual_comparison}
\end{figure*}

\textbf{Qualitative Comparison:} Fig.~\ref{fig:Visual_comparison} presents representative visual comparisons across multiple target domains. Compared with existing PEFT methods, FAN-LoRA consistently produces segmentation maps with sharper boundaries and more anatomically consistent structures.

Specifically, linear adapters such as LoRA and DeLoRA tend to generate over-smoothed predictions, exhibiting limitations in consistently recovering fine boundary details under strong domain shifts. FourierFT partially improves high-frequency detail preservation but often introduces spurious noise or oscillatory artifacts due to the lack of structural constraints. AuroRA improves global smoothness via non-linear mapping; however, it still exhibits blurred boundaries in challenging regions, indicating insufficient high-frequency compensation.

In contrast, FAN-LoRA achieves a superior balance between structural coherence and detail preservation. The low-pass ANL branch effectively aligns global anatomical structures, while the high-pass Fourier branch selectively enhances boundary sharpness and local textures. This frequency-decoupled design leads to more precise delineation of organ boundaries and better recovery of thin or low-contrast structures, validating the effectiveness of the proposed approach. Quantitative boundary precision is further validated by the lowest HD95 across all target domains (see Tables~\ref{tab:chaos_comparison}, \ref{tab:prostate_comparison} and \ref{tab:mmwhs_comparison}).

\subsection{Ablation Study}
\textbf{Ablation Study on Core Modules:} Table~\ref{tab:ablation_module} provides empirical evidence supporting our core hypothesis. The independent introduction of either the low-pass or high-pass branch yields only marginal improvements in average Dice; however, the high-pass branch alone already reduces HD95 from 30.51 to 30.19 on the 1.5T domain and from 28.41 to 27.13 on the 3T domain, demonstrating that explicit high-frequency modeling provides independent boundary-refinement benefits even in the absence of structural guidance, indicating that neither structural alignment nor detail compensation alone is sufficient. The full FAN-LoRA model synergizes both branches to achieve the highest performance. This confirms our hypothesis that global structural alignment (low-pass) and local detail compensation (high-pass) are highly complementary and indispensable for robust domain adaptation. 

\begin{table}[ht]
\centering
\caption{Ablation study of the core low-pass and high-pass modules in FAN-LoRA on the NCI-ISBI target domain}
\label{tab:ablation_module}
\resizebox{1.0\linewidth}{!}{
\begin{tabular}{ccc|cc|cc}
\toprule
\multirow{2}{*}{Base LoRA} & \multirow{2}{*}{Low-pass} & \multirow{2}{*}{High-pass} & \multicolumn{2}{c|}{NCI-ISBI 1.5T} & \multicolumn{2}{c}{NCI-ISBI 3T}  \\
 &  &  &  Avg. & HD95 $\downarrow$ & Avg. & HD95 $\downarrow$ \\
\midrule
\checkmark & & & 78.6 & 30.51 & 78.1 & 28.41 \\
\checkmark & \checkmark & & 79.1 & 29.08 & 78.6 & 28.19 \\
\checkmark & & \checkmark & 79.5 & 30.19 & 78.7 & 27.13 \\
\checkmark & \checkmark & \checkmark & \textbf{80.9} & \textbf{26.96} & \textbf{79.6} & \textbf{26.48} \\
\bottomrule
\end{tabular}
}
\end{table}

\begin{table}[ht]
\centering
\caption{Ablation study on high-pass branch components on the NCI-ISBI target domain.}
\label{tab:highpass_ablation}
\resizebox{1.0\linewidth}{!}{
\begin{tabular}{c c|c c|c c}
\toprule
\multirow{2}{*}{$U_\text{high}, V_\text{high}$} & \multirow{2}{*}{$S_\text{freq}$} & \multicolumn{2}{c|}{NCI-ISBI 1.5T} & \multicolumn{2}{c}{NCI-ISBI 3T} \\
 &  &  Avg. & HD95 $\downarrow$ & Avg. & HD95 $\downarrow$ \\
\midrule
Random \& Frozen       &  & 75.9  & 36.09 & 77.3 & 30.63 \\
Learnable              &  & 78.9  & 35.25 & 78.05 & 28.75 \\
Fixed Sinusoidal       &  & 79.6 & 28.63 & 78.9 & 27.11 \\
Random \& Frozen       & \checkmark     & 77.3 & 37.24 & 78.3 & 29.36 \\
Learnable              & \checkmark     & 79.2  & 30.58 & 78.6 & 28.97 \\
Fixed Sinusoidal       & \checkmark    & \textbf{80.9} & \textbf{26.96} & \textbf{79.6} & \textbf{26.48} \\
\bottomrule
\end{tabular}
}
\end{table}

We further dissect the high-pass branch by ablating its two key components: the projection matrices $U_\text{high}, V_\text{high}$ and the frequency modulation scalar $S_\text{freq}$. Table~\ref{tab:highpass_ablation} reports the results under different configurations. Without the modulation scalar, fixed sinusoidal projections substantially outperform randomly frozen and learnable matrices, reducing HD95 from over 35 to 28.63 on the 1.5T domain, which confirms that a structured frequency basis is essential for boundary refinement. The learnable scalar $S_\text{freq}$ degrades performance when paired with random projections, yet it brings a clear gain when combined with the sinusoidal basis, achieving the best HD95 of 26.96 on 1.5T and 26.48 on 3T. We therefore adopt fixed sinusoidal projections with a learnable $S_\text{freq}$ as the default high-pass design.

\textbf{Hyperparameter Sensitivity:} Table~\ref{tab:prostate_hyperparameters} details the sensitivity analysis regarding the intrinsic rank of the low-pass space and the number of sampled frequencies $n$ in the high-pass branch. Our results reveal that an optimal balance is struck at a lower rank dimension combined with a concise frequency sampling, achieving peak Dice scores on both 1.5T and 3T targets. Increasing the rank or the frequency dimensions excessively leads to marginal performance degradation, likely due to the over-parameterization of the low-rank subspace which inadvertently re-introduces high-frequency noise fitting. Furthermore, this phenomenon suggests that within the pre-trained MedSAM feature space, the intrinsic rank of the global anatomical transformations between medical modalities is inherently low, making higher rank dimensions redundant and prone to overfitting.

Regarding the high-frequency sampling count $n$, the optimal value of $n=16$ emerges from a trade-off between representation capacity and over-parameterization. The high-frequency projection matrices $\mathbf{U}_\text{high}$ and $\mathbf{V}_\text{high}$ are constructed via discrete sinusoidal sampling along integer feature dimensions. As the sampling density increases, the column vectors of these projection matrices become increasingly correlated, reducing the effective rank of the projected spectral subspace. Selecting $n \ll d_\text{in}$ ensures a well-conditioned basis for high-frequency extraction. Empirically, $n=16$ strikes an optimal balance: smaller values (e.g., $n=8$) under-represent the high-frequency subspace, limiting texture compensation; larger values (e.g., $n \geq 64$) introduce redundant spectral components that increase overfitting risk without measurable performance gains. We recommend $n$ as a lightweight hyperparameter amenable to minimal grid search in the range $[8, 64]$, with the understanding that cross-modality shifts may benefit from slightly larger $n$ than cross-center shifts due to stronger high-frequency perturbations.

\begin{table}[ht]
\centering
\caption{Hyperparameter sensitivity analysis of rank dimension and frequency sampling on the NCI-ISBI target domain}
\label{tab:prostate_hyperparameters}
\resizebox{1.0\linewidth}{!}{
\begin{tabular}{l|l|cccc|cccc}
\toprule
Rank & Frequency & \multicolumn{4}{c|}{NCI-ISBI 1.5T} & \multicolumn{4}{c}{NCI-ISBI 3T} \\
& & PZ & CG & Avg. & HD95 $\downarrow$ & PZ & CG & Avg. & HD95 $\downarrow$ \\
\midrule
4  & \multirow{4}{*}{16} & 66.2 & 92.2 & 79.2 & 30.38 & 65.1 & 92.9 & 79 & 27.71  \\
8  &                     & 66.1 & 93.8 & 80 & 30.18 & 59.6 & 93.2 & 76.4 & 32.77 \\
16 &                     & 65.7 & 92.9 & 79.3  & 30.68 & 61.2 & \textbf{93.4} & 77.3 & 31.42 \\
32 &                     & 65.8 & 93.2 & 79.5  & 33.44 & 59   & 93.3 & 76.15 & 33.47 \\
\midrule
\multirow{4}{*}{2}  & 32  & 66.7 & 93.2 & 80 & 29.57 & 59.3 & 92.6 & 76 & 32.15 \\
& 64  & 66.6 & 92.6 & 79.6  & 29.7 & 63.9 & 92.4 & 78.2 & 28.12 \\
& 128 & 66.6 & \textbf{93.5} & 80.1 & 29.32 & 60.9 & 92.4 & 76.7 & 29.72 \\
& 256 & 67.1 & 92.9 & 80    & 29.35 & 60.1 & 92.5 & 76.3 & 31.05 \\
\midrule
2 & 16 & \textbf{68.4}& 93.3& \textbf{80.9} & \textbf{26.96} & \textbf{66.4}& 92.8 & \textbf{79.6} & \textbf{26.48} \\
\bottomrule
\end{tabular}
}
\end{table}

\textbf{Error Analysis:} Despite its competitive performance, FAN-LoRA may still struggle in extremely low-contrast regions where both structural and textural cues are weak. In such cases, the high-frequency branch may have limited signal to enhance, while the low-pass branch may over-smooth ambiguous boundaries.

\section{Conclusion}
In this paper, we presented FAN-LoRA, a frequency-decoupled PEFT framework that explicitly separates low- and high-frequency optimization for medical foundation model adaptation. Extensive experiments demonstrate competitive advantages over existing PEFT baselines in both accuracy and boundary precision. While highly effective, we note that the performance gains may diminish in extremely low-contrast regions where frequency cues are intrinsically weak. Future work will explore adaptive frequency-band selection and theoretically grounded gradient decomposition for broader multi-domain generalization. 

{
    \bibliographystyle{IEEEtran}
    \bibliography{main}

@inproceedings{kirillov2023segment,
  title={Segment anything},
  author={Kirillov, Alexander and Mintun, Eric and Ravi, Nikhila and Mao, Hanzi and Rolland, Chloe and Gustafson, Laura and Xiao, Tete and Whitehead, Spencer and Berg, Alexander C and Lo, Wan-Yen and others},
  booktitle={Proceedings of the IEEE/CVF international conference on computer vision},
  pages={4015--4026},
  year={2023}
}

@article{MedSAM,
  title={Segment Anything in Medical Images},
  author={Ma, Jun and He, Yuting and Li, Feifei and Han, Lin and You, Chenyu and Wang, Bo},
  journal={Nature Communications},
  volume={15},
  pages={654},
  year={2024}
}

@article{wu2025medical,
  title={Medical sam adapter: Adapting segment anything model for medical image segmentation},
  author={Wu, Junde and Wang, Ziyue and Hong, Mingxuan and Ji, Wei and Fu, Huazhu and Xu, Yanwu and Xu, Min and Jin, Yueming},
  journal={Medical image analysis},
  volume={102},
  pages={103547},
  year={2025},
  publisher={Elsevier}
}

@article{hu2022lora,
  title={Lora: Low-rank adaptation of large language models.},
  author={Hu, Edward J and Shen, Yelong and Wallis, Phillip and Allen-Zhu, Zeyuan and Li, Yuanzhi and Wang, Shean and Wang, Liang and Chen, Weizhu and others},
  journal={Iclr},
  volume={1},
  number={2},
  pages={3},
  year={2022}
}

@article{samed,
  title={Customized Segment Anything Model for Medical Image Segmentation},
  author={Kaidong Zhang and Dong Liu},
  journal={arXiv preprint arXiv:2304.13785},
  year={2023}
}

@article{dong2025aurora,
  title={Aurora: Breaking low-rank bottleneck of lora with nonlinear mapping},
  author={Dong, Haonan and Zhu, Wenhao and Song, Guojie and Wang, Liang},
  journal={arXiv preprint arXiv:2505.18738},
  year={2025}
}

@inproceedings{gao2024parameter,
  title={Parameter-Efficient Fine-Tuning with Discrete Fourier Transform},
  author={Gao, Ziqi and Wang, Qichao and Chen, Aochuan and Liu, Zijing and Wu, Bingzhe and Chen, Liang and Li, Jia},
  booktitle={International Conference on Machine Learning},
  pages={14884--14901},
  year={2024},
  organization={PMLR}
}

@article{he2025adaptfrcnet,
  title={AdaptFRCNet: Semi-supervised adaptation of pre-trained model with frequency and region consistency for medical image segmentation},
  author={He, Along and Wu, Yanlin and Wang, Zhihong and Li, Tao and Fu, Huazhu},
  journal={Medical Image Analysis},
  volume={103},
  pages={103626},
  year={2025},
  publisher={Elsevier}
}

@inproceedings{ly2025frequency,
  title={Frequency Strikes Back: Boosting Parameter-Efficient Foundation Model Adaptation for Medical Imaging},
  author={Ly, Son T and Nguyen, Hien V},
  booktitle={International Conference on Medical Image Computing and Computer-Assisted Intervention},
  pages={258--267},
  year={2025},
  organization={Springer}
}

@article{bini2025delora,
  title={Delora: Decoupling angles and strength in low-rank adaptation},
  author={Bini, Massimo and Girrbach, Leander and Akata, Zeynep},
  journal={arXiv preprint arXiv:2503.18225},
  year={2025}
}

@article{bhardwaj2024shira,
  title={Rapid switching and multi-adapter fusion via sparse high rank adapters},
  author={Bhardwaj, Kartikeya and Pandey, Nilesh Prasad and Priyadarshi, Sweta and Ganapathy, Viswanath and Esteves, Rafael and Kadambi, Shreya and Borse, Shubhankar and Whatmough, Paul and Garrepalli, Risheek and Van Baalen, Mart and others},
  journal={arXiv preprint arXiv:2407.16712},
  year={2024}
}

@article{isensee2021nnu,
  title={nnU-Net: a self-configuring method for deep learning-based biomedical image segmentation},
  author={Isensee, Fabian and Jaeger, Paul F and Kohl, Simon AA and Petersen, Jens and Maier-Hein, Klaus H},
  journal={Nature methods},
  volume={18},
  number={2},
  pages={203--211},
  year={2021},
  publisher={Nature Publishing Group US New York}
}

@inproceedings{yang2020fda,
  title={Fda: Fourier domain adaptation for semantic segmentation},
  author={Yang, Yanchao and Soatto, Stefano},
  booktitle={Proceedings of the IEEE/CVF conference on computer vision and pattern recognition},
  pages={4085--4095},
  year={2020}
}

@article{mazurowski2023segment,
  title={Segment anything model for medical image analysis: an experimental study},
  author={Mazurowski, Maciej A and Dong, Haoyu and Gu, Hanxue and Yang, Jichen and Konz, Nicholas and Zhang, Yixin},
  journal={Medical Image Analysis},
  volume={89},
  pages={102918},
  year={2023},
  publisher={Elsevier}
}

@article{huang2024segment,
  title={Segment anything model for medical images?},
  author={Huang, Yuhao and Yang, Xin and Liu, Lian and Zhou, Han and Chang, Ao and Zhou, Xinrui and Chen, Rusi and Yu, Junxuan and Chen, Jiongquan and Chen, Chaoyu and others},
  journal={Medical Image Analysis},
  volume={92},
  pages={103061},
  year={2024},
  publisher={Elsevier}
}

@misc{ji2024segment,
  title={Segment anything is not always perfect: An investigation of sam on different real-world applications},
  author={Ji, Wei and Li, Jingjing and Bi, Qi and Liu, Tingwei and Li, Wenbo and Cheng, Li},
  year={2024},
  publisher={Springer}
}

@inproceedings{yao2024fnpc,
  title={FNPC-SAM: uncertainty-guided false negative/positive control for SAM on noisy medical images},
  author={Yao, Xing and Liu, Han and Hu, Dewei and Lu, Daiwei and Lou, Ange and Li, Hao and Deng, Ruining and Arenas, Gabriel and Oguz, Baris and Schwartz, Nadav and others},
  booktitle={Medical Imaging 2024: Image Processing},
  volume={12926},
  pages={1292602},
  year={2024},
  organization={SPIE}
}

@article{phuntsho2025adaptation,
  title={Adaptation of Foundation Models for Medical Image Analysis: Strategies, Challenges, and Future Directions},
  author={Phuntsho, Karma and Lee, Kyungmi and Lee, Ickjai and Ahn, Euijoon and others},
  journal={arXiv preprint arXiv:2511.01284},
  year={2025}
}

@article{silva2025towards,
  title={Towards foundation models and few-shot parameter-efficient fine-tuning for volumetric organ segmentation},
  author={Silva-Rodr{\'\i}guez, Julio and Dolz, Jose and Ayed, Ismail Ben},
  journal={Medical Image Analysis},
  volume={103},
  pages={103596},
  year={2025},
  publisher={Elsevier}
}

@inproceedings{houlsby2019parameter,
  title={Parameter-efficient transfer learning for NLP},
  author={Houlsby, Neil and Giurgiu, Andrei and Jastrzebski, Stanislaw and Morrone, Bruna and De Laroussilhe, Quentin and Gesmundo, Andrea and Attariyan, Mona and Gelly, Sylvain},
  booktitle={International conference on machine learning},
  pages={2790--2799},
  year={2019},
  organization={PMLR}
}

@inproceedings{cai2025loki,
  title={Loki: Low-dimensional kan for efficient fine-tuning image models},
  author={Cai, Xuan and Pan, Renjie and Yang, Hua},
  booktitle={Proceedings of the Computer Vision and Pattern Recognition Conference},
  pages={14869--14880},
  year={2025}
}

@article{li2025adaptive,
  title={Adaptive Frequency Domain Alignment Network for Medical image segmentation},
  author={Li, Zhanwei and Li, Liang and Zhang, Jiawan},
  journal={arXiv preprint arXiv:2512.16393},
  year={2025}
}

@inproceedings{jiang2025sr,
  title={SR-SAM: Subspace Regularization for Domain Generalization of Segment Anything Model},
  author={Jiang, Xixi and Yang, Chen and Zhang, Liang and Cheng, Tim Kwang-Ting and Yang, Xin},
  booktitle={International Conference on Medical Image Computing and Computer-Assisted Intervention},
  pages={510--520},
  year={2025},
  organization={Springer}
}

@article{lialin2023scaling,
  title={Scaling down to scale up: A guide to parameter-efficient fine-tuning},
  author={Lialin, Vladislav and Deshpande, Vijeta and Yao, Xiaowei and Rumshisky, Anna},
  journal={arXiv preprint arXiv:2303.15647},
  year={2023}
}

@article{chen2022adaptformer,
  title={Adaptformer: Adapting vision transformers for scalable visual recognition},
  author={Chen, Shoufa and Ge, Chongjian and Tong, Zhan and Wang, Jiangliu and Song, Yibing and Wang, Jue and Luo, Ping},
  journal={Advances in Neural Information Processing Systems},
  volume={35},
  pages={16664--16678},
  year={2022}
}

@inproceedings{liu2021feddg,
  title={Feddg: Federated domain generalization on medical image segmentation via episodic learning in continuous frequency space},
  author={Liu, Quande and Chen, Cheng and Qin, Jing and Dou, Qi and Heng, Pheng-Ann},
  booktitle={Proceedings of the IEEE/CVF conference on computer vision and pattern recognition},
  pages={1013--1023},
  year={2021}
}

@article{bougourzi2025recent,
  title={Recent advances in medical imaging segmentation: A survey},
  author={Bougourzi, Fares and Hadid, Abdenour},
  journal={arXiv preprint arXiv:2505.09274},
  year={2025}
}

@article{wang2022medical,
  title={Medical image segmentation using deep learning: A survey},
  author={Wang, Risheng and Lei, Tao and Cui, Ruixia and Zhang, Bingtao and Meng, Hongying and Nandi, Asoke K},
  journal={IET image processing},
  volume={16},
  number={5},
  pages={1243--1267},
  year={2022},
  publisher={Wiley Online Library}
}

@book{ma2023flare,
  title={Fast and Low-Resource Semi-supervised Abdominal Organ Segmentation: MICCAI 2022 Challenge, FLARE 2022, Held in Conjunction with MICCAI 2022, Singapore, September 22, 2022, Proceedings},
  author={Ma, Jun and Wang, Bo},
  year={2023},
  publisher={Springer Nature}
}

@article{kavur2021chaos,
  title={CHAOS challenge-combined (CT-MR) healthy abdominal organ segmentation},
  author={Kavur, A Emre and Gezer, N Sinem and Bar{\i}{\c{s}}, Mustafa and Aslan, Sinem and Conze, Pierre-Henri and Groza, Vladimir and Pham, Duc Duy and Chatterjee, Soumick and Ernst, Philipp and {\"O}zkan, Sava{\c{s}} and others},
  journal={Medical image analysis},
  volume={69},
  pages={101950},
  year={2021},
  publisher={Elsevier}
}

@article{zhuang2019mmwhs,
  title={Evaluation of algorithms for multi-modality whole heart segmentation: an open-access grand challenge},
  author={Zhuang, Xiahai and Li, Lei and Payer, Christian and {\v{S}}tern, Darko and Urschler, Martin and Heinrich, Mattias P and Oster, Julien and Wang, Chunliang and Smedby, {\"O}rjan and Bian, Cheng and others},
  journal={Medical image analysis},
  volume={58},
  pages={101537},
  year={2019},
  publisher={Elsevier}
}

@article{litjens2014promise,
  title={Evaluation of prostate segmentation algorithms for MRI: the PROMISE12 challenge},
  author={Litjens, Geert and Toth, Robert and Van De Ven, Wendy and Hoeks, Caroline and Kerkstra, Sjoerd and Van Ginneken, Bram and Vincent, Graham and Guillard, Gwenael and Birbeck, Neil and Zhang, Jindang and others},
  journal={Medical image analysis},
  volume={18},
  number={2},
  pages={359--373},
  year={2014},
  publisher={Elsevier}
}

@dataset{bloch2015nciisbi,
  author    = {Bloch, Nicholas and Madabhushi, Anant and Huisman, Henkjan and
               Freymann, John and Kirby, Justin and Grauer, Michael and
               Enquobahrie, Andinet and Jaffe, Carl and Clarke, Larry and Farahani, Keyvan},
  title     = {{NCI}-{ISBI} 2013 Challenge: Automated Segmentation of Prostate Structures},
  year      = {2015},
  publisher = {The Cancer Imaging Archive},
  doi       = {10.7937/K9/TCIA.2015.zF0vlOPv}
}

@article{cordero2020generalized,
  title={Generalized Born-Jordan distributions and applications},
  author={Cordero, Elena and de Gosson, Maurice and D{\"o}rfler, Monika and Nicola, Fabio},
  journal={Advances in Computational Mathematics},
  volume={46},
  number={4},
  pages={51},
  year={2020},
  publisher={Springer}
}

@article{unser1993family,
  title={A family of polynomial spline wavelet transforms},
  author={Unser, Michael and Aldroubi, Akram and Eden, Murray},
  journal={Signal processing},
  volume={30},
  number={2},
  pages={141--162},
  year={1993},
  publisher={Elsevier}
}

@article{panda2001least,
  title={Least squares generalized B-spline signal and image processing},
  author={Panda, Rutuparna and Chatterji, BN},
  journal={Signal Processing},
  volume={81},
  number={10},
  pages={2005--2017},
  year={2001},
  publisher={Elsevier}
}
}

\end{document}